\documentclass[preprint]{elsarticle}

\usepackage[spanish,es-nodecimaldot]{babel}
\usepackage[utf8x]{inputenc}
\usepackage[T1]{fontenc} 
\usepackage{lmodern}
\usepackage{amsmath}
\usepackage{amsfonts}
\usepackage{amssymb}
\usepackage{graphicx}
\usepackage{wrapfig}
\usepackage{pstricks,pst-all,pst-circ,pst-3dplot,pst-plot}
\usepackage{framed}
\usepackage[hyperref,framed]{ntheorem}
\usepackage{array}
\usepackage{verbatim}
\usepackage{fancyvrb}
\usepackage{ifthen}
\usepackage{calc}
\usepackage{lscape}
\usepackage{rotating}
\usepackage{url}
\usepackage{enumerate}
\usepackage[subrefformat=parens,labelformat=parens]{subfig}
\usepackage{rotating}
\usepackage{colortbl}
\usepackage[boxed,linesnumbered]{algorithm2e}

\title{A family of OWA operators based on Faulhaber's formulas}

\author[unc]{O.G.~Duarte\corref{cor1}}
\ead{ogduartev@unal.edu.co}

\author[unc]{S.M.~Téllez\corref{cor2}}
\ead{smtellezg@unal.edu.co}

\address[unc]{Universidad Nacional de Colombia, Facultad de Ingeniería, Bogotá, Colombia}

\cortext[cor1]{Corresponding author}
\cortext[cor2]{This work was partially supported by Colciencias, call number 617}

\newcommand{\plotoptionslinedefault}
{
  \psset{linecolor=black}
}

\newcommand{\plotoptionsline}[1]
{
\plotoptionslinedefault{}
\ifthenelse{\equal{#1}{1}}{\psset{linecolor=black}}{}
\ifthenelse{\equal{#1}{2}}{\psset{linecolor=blue}}{}
\ifthenelse{\equal{#1}{3}}{\psset{linecolor=red}}{}
\ifthenelse{\equal{#1}{4}}{\psset{linecolor=cyan}}{}
\ifthenelse{\equal{#1}{5}}{\psset{linecolor=magenta}}{}
\ifthenelse{\equal{#1}{6}}{\psset{linecolor=gray}}{}
}

\newcommand{\plotoptionsbarradefault}
{
  \psset{linecolor=black,fillcolor=white,fillstyle=none}
}

\newcommand{\plotoptionsbarra}[1]
{
\plotoptionsbarradefault{}
\plotoptionsline{#1}
\ifthenelse{\equal{#1}{1}}{\psset{fillstyle=solid}}{}
\ifthenelse{\equal{#1}{2}}{\psset{fillstyle=solid}}{}
\ifthenelse{\equal{#1}{3}}{\psset{fillstyle=solid}}{}
\ifthenelse{\equal{#1}{4}}{\psset{fillstyle=solid}}{}
\ifthenelse{\equal{#1}{5}}{\psset{fillstyle=solid}}{}
\ifthenelse{\equal{#1}{6}}{\psset{fillstyle=solid}}{}
}

\newcommand{\plotpeso}[2]
{
  \plotoptionsline{#2}
  \fileplot{figs/data/P#1-W#2.dat}
}

\newlength{\dyLabel}
\newcommand{\plotlabel}[1]
{
  \plotoptionsline{#1}
  \rput[l](1.05,1.1)
  {
    \psline(0,-#1\dyLabel)(0.1,-#1\dyLabel)
    \rput[l](0.11,-#1\dyLabel){$w_{#1}$}
  }
}

\newcommand{\plotlabelvariable}[3]
{
  \plotoptionsline{#1}
  \rput[lt](1.05,2.5cm)
  {
    \psline(0,-#2\dyLabel)(0.1,-#2\dyLabel)
    \rput[l](0.11,-#2\dyLabel){#3}
  }
}

\newlength{\dyLabelBarra}
\newcommand{\plotlabelbarras}[3]
{
  \plotoptionsline{#1}
  \rput[lt](6.05,4cm)
  {
    \psline(0.5,-#2\dyLabelBarra)(1,-#2\dyLabelBarra)
    \rput[l](1.11,-#2\dyLabelBarra){#3}
  }
}

\newcommand{\plotcaso}[1]
{
  \multido{\i=1+1}{5}
  {
    \plotpeso{#1}{\i}
    \plotlabel{\i}
  }
}

\newcommand{\plotvariable}[2]
{
  \ifthenelse{\equal{#2}{A}}
  {
    \multido{\i=1+1}{6}
    {
      \plotoptionsline{\i}
      \fileplot[]{figs/data/P\i-#1.dat}
      \plotlabelvariable{\i}{\i}{\labelcaso{\i}}
    }
  }{}
  
  \ifthenelse{\equal{#2}{B}}
  {
    \plotoptionsline{1}\fileplot[]{figs/data/P1-#1.dat}\plotlabelvariable{1}{2}{\labelcaso{1}}
    \plotoptionsline{2}\fileplot[]{figs/data/P2-#1.dat}\plotlabelvariable{2}{3}{\labelcaso{2}}
    \plotoptionsline{3}\fileplot[]{figs/data/P3-#1.dat}\plotlabelvariable{3}{4}{\labelcaso{3}}
  }{}
  
  \ifthenelse{\equal{#2}{C}}
  {
    \plotoptionsline{4}\fileplot[]{figs/data/P4-#1.dat}\plotlabelvariable{4}{2}{\labelcaso{4}}
    \plotoptionsline{5}\fileplot[]{figs/data/P5-#1.dat}\plotlabelvariable{5}{3}{\labelcaso{5}}
    \plotoptionsline{6}\fileplot[]{figs/data/P6-#1.dat}\plotlabelvariable{6}{4}{\labelcaso{6}}
  }{}
  
  \ifthenelse{\equal{#2}{D}}
  {
    \plotoptionsline{1}\fileplot[]{figs/data/P1-#1.dat}\plotlabelvariable{1}{1}{\labelcaso{1}}
    \plotoptionsline{2}\fileplot[]{figs/data/P2-#1.dat}\plotlabelvariable{2}{2}{\labelcaso{2}}
    \plotoptionsline{3}\fileplot[]{figs/data/P6-#1.dat}\plotlabelvariable{3}{3}{\labelcaso{6}}
  }{}
}

\newcommand{\labelcaso}[1]
{%
\ifthenelse{\equal{#1}{1}}{Maximum entropy}{}%
\ifthenelse{\equal{#1}{2}}{Exponential with preset}{}%
\ifthenelse{\equal{#1}{3}}{Exponential without preset}{}%
\ifthenelse{\equal{#1}{4}}{Linear with $\beta=1.00$}{}%
\ifthenelse{\equal{#1}{5}}{Linear with $\beta=1.25$}{}%
\ifthenelse{\equal{#1}{6}}{Linear with $\beta=1.50$}{}%
}

\newlength{\dBarra}
\newlength{\wBarra}

\newcommand{\plotpuntobarra}[2]
{
  \setlength{\wBarra}{#1cm}
  \addtolength{\wBarra}{\dBarra}
  \addtolength{\wBarra}{\dBarra}
  \psline(#1,0)(#1,#2)
  \rput[c](#1,#2){\pscircle{\dBarra}}
}

\newcommand{\plotbarraline}[1] 
{
  \pscurve[fillstyle=none,linestyle=dashed,showpoints=false](NW#1-2)(NW#1-3)(NW#1-4)(NW#1-5)
}

\newcommand{\plotbarras}[1]
{
  \ifthenelse{\equal{#1}{A}}
  {
    \plotoptionsbarra{1}
    \rput[bl](-1\dBarra,0){\plotpuntobarra{1}{0.45935872165273}
\pnode(1,0.45935872165273){NW1-1}
\plotpuntobarra{2}{0.26079338775684}
\pnode(2,0.26079338775684){NW1-2}
\plotpuntobarra{3}{0.14806199275457}
\pnode(3,0.14806199275457){NW1-3}
\plotpuntobarra{4}{0.084060702931119}
\pnode(4,0.084060702931119){NW1-4}
\plotpuntobarra{5}{0.047724882219085}
\pnode(5,0.047724882219085){NW1-5}
}
    \plotbarraline{1}
    \plotlabelbarras{1}{1}{\labelcaso{1}}

    \plotoptionsbarra{2}
    \rput[bl](1\dBarra,0){\plotpuntobarra{1}{0.62369956929707}
\pnode(1,0.62369956929707){NW2-1}
\plotpuntobarra{2}{0.078129649725687}
\pnode(2,0.078129649725687){NW2-2}
\plotpuntobarra{3}{0.087916968454496}
\pnode(3,0.087916968454496){NW2-3}
\plotpuntobarra{4}{0.09893036786597}
\pnode(4,0.09893036786597){NW2-4}
\plotpuntobarra{5}{0.11132344465678}
\pnode(5,0.11132344465678){NW2-5}
}
    \plotbarraline{2}
    \plotlabelbarras{2}{2}{\labelcaso{2}}
  }{}
  
  \ifthenelse{\equal{#1}{B}}
  {
    \plotoptionsbarra{4}
    \rput[bl](-3\dBarra,0){\plotpuntobarra{1}{0.6}
\pnode(1,0.6){NW4-1}
\plotpuntobarra{2}{0.1}
\pnode(2,0.1){NW4-2}
\plotpuntobarra{3}{0.1}
\pnode(3,0.1){NW4-3}
\plotpuntobarra{4}{0.1}
\pnode(4,0.1){NW4-4}
\plotpuntobarra{5}{0.1}
\pnode(5,0.1){NW4-5}
}
    \plotbarraline{4}
    \plotlabelbarras{4}{1}{\labelcaso{4}}

    \plotoptionsbarra{5}
    \rput[bl](-1\dBarra,0){\plotpuntobarra{1}{0.5363591176868}
\pnode(1,0.5363591176868){NW5-1}
\plotpuntobarra{2}{0.1636408823132}
\pnode(2,0.1636408823132){NW5-2}
\plotpuntobarra{3}{0.1318204411566}
\pnode(3,0.1318204411566){NW5-3}
\plotpuntobarra{4}{0.1}
\pnode(4,0.1){NW5-4}
\plotpuntobarra{5}{0.068179558843399}
\pnode(5,0.068179558843399){NW5-5}
}
    \plotbarraline{5}
    \plotlabelbarras{5}{2}{\labelcaso{5}}

    \plotoptionsbarra{6}
    \rput[bl](1\dBarra,0){\plotpuntobarra{1}{0.48284382523312}
\pnode(1,0.48284382523312){NW6-1}
\plotpuntobarra{2}{0.21715617476688}
\pnode(2,0.21715617476688){NW6-2}
\plotpuntobarra{3}{0.15857808738344}
\pnode(3,0.15857808738344){NW6-3}
\plotpuntobarra{4}{0.1}
\pnode(4,0.1){NW6-4}
\plotpuntobarra{5}{0.041421912616562}
\pnode(5,0.041421912616562){NW6-5}
}
    \plotbarraline{6}
    \plotlabelbarras{6}{3}{\labelcaso{6}}
  }{}
    
  \ifthenelse{\equal{#1}{D}}
  {
  }{}
  
  \ifthenelse{\equal{#1}{E}}
  {
  }{}
  
}

\newcommand{\weightVsOrness}[1]
{
 \begin{pspicture}(0,0)(6,3.0)
  \rput[bl](1,0.5)
  {
    \psset{xunit=4}
    \psset{yunit=2}
    \scriptsize
    \psaxes[Dx=0.2,Dy=0.25,axesstyle=frame,tickstyle=top]{->}(0,0)(1.0001,1.0)[$Orness$,0][$Weight$,90]
    \plotcaso{#1}
  }
 \end{pspicture}
}

\newcommand{\variableVsOrness}[5]
{
 \begin{pspicture}(0,0)(9,3.0)
  \rput[bl](1,0.5)
  {
    \psset{xunit=4}
    \psset{yunit=#4\psyunit}
    \scriptsize
    \psaxes[Dx=0.2,Dy=0.25,axesstyle=frame,tickstyle=top]{->}(0,0)(1.0001,#3)[{\it Orness},0][{\it #2},90]
    \plotvariable{#1}{#5}
  }
 \end{pspicture}
}

\newcommand{\variableVsOrnessNcien}[4]
{
 \begin{pspicture}(0,0)(6,5.5)
  \rput[bl](1,1)
  {
    \psset{xunit=40}
    \psset{yunit=#4\psyunit}
    \scriptsize
    \rput[bl](-0.9,0)
    {
     \psaxes[Ox=0.9,Dx=0.02,Dy=0.5,axesstyle=frame,tickstyle=top,labels=y]%
            {->}(0.9,0)(0.9,0)(1.0001,#3)[{\it Orness},0][{\it #2},90]
     \multido{\i=0+2}{5}{\rput[t](0.9\i,-0.2){$0.9\i$}}
     \rput[t](1.0,-0.2){1.0}
      \plotoptionsline{1}\fileplot[]{figs/data/PN100-#1.dat}
    }
  }
 \end{pspicture}
}

\newcommand{\plotweights}[2]
{
 \begin{pspicture}(0,0)(13,6)
  \rput[bl](1,1)
  {
    \multido{\i=1+1}{5}{\rput[t](\i,-0.2){$w_\i$}}
    \psset{xunit=1}
    \psset{yunit=4}
    \psaxes[Dx=1,Dy=0.2,axesstyle=frame,tickstyle=top,labels=y,ticks=y]{->}(0,0)(6,1.1)[{},0][{\it Weight},90]
    \plotbarras{#2}
  }
 \end{pspicture}
}

\newcommand{\owasSimple}[1]
{
 \begin{pspicture}(0,0)(4.5,4)
  \rput[bl](0.5,1)
  {
    \psset{xunit=0.75}
    \psset{yunit=2}
    \multido{\i=1+1}{5}{\rput[t](\i,-0.2){$w_\i$}}
    \rput[r](-0.1,1){1.0}
    \rput[r](-0.1,0.2){$\frac{1}{n}$}
    \psaxes[Dx=1,Dy=1,axesstyle=frame,tickstyle=top,labels=none,ticks=none]{->}(0,0)(5.5,1.1)[{},0][{\it Weight},90]
    \plotoptionsbarra{1}
    \ifthenelse{\equal{#1}{1}}
    {
      \psline[linestyle=dotted](0,1)(5.0,1)
      \plotpuntobarra{1}{0}
      \plotpuntobarra{2}{0}
      \plotpuntobarra{3}{0}
      \plotpuntobarra{4}{0}
      \plotpuntobarra{5}{1}
    }{}
    \ifthenelse{\equal{#1}{2}}
    {
      \psline[linestyle=dotted](0,0.2)(5.0,0.2)
      \plotpuntobarra{1}{0.2}
      \plotpuntobarra{2}{0.2}
      \plotpuntobarra{3}{0.2}
      \plotpuntobarra{4}{0.2}
      \plotpuntobarra{5}{0.2}
    }{}
    \ifthenelse{\equal{#1}{3}}
    {
      \psline[linestyle=dotted](0,1)(1,1)
      \plotpuntobarra{1}{1}
      \plotpuntobarra{2}{0}
      \plotpuntobarra{3}{0}
      \plotpuntobarra{4}{0}
      \plotpuntobarra{5}{0}
    }{}
  }
 \end{pspicture}
}

\begin{document}
\begin{abstract}
 In this paper we develop a new family of Ordered Weighted Averaging (OWA) operators. Weight vector is obtained from a desired orness of the operator. Using Faulhaber's formulas we obtain direct and simple expressions for the weight vector without any iteration loop. With the exception of one weight, the remaining follow a straight line relation. As a result, a fast and robust algorithm is developed. The resulting weight vector is suboptimal according with the Maximum Entropy criterion, but it is very close to the optimal. Comparisons are done with other procedures. 
\end{abstract}


\maketitle

\section{Introduction}

Ordered Weighted Averaging (OWA) operators are well known aggregation operators \cite{Detyniecki2001} that have received great attention in recent years. Although OWA operators were proposed by Yager since 1988 \cite{Yager1998}, a great amount of research arises since 2008. \cite{Yager2011} provides a collection of recent developments around OWA operators. A systematic review of literature about OWA is done in \cite{Marra2014}; it reveals some great research lines, one of them is to solve the weight vector determination problem. Another wide area of research deals with generalizations to fuzzy set theory.

The weight vector determines the way in which the OWA operator aggregates information. For some applications we need to adjust the behaivor of the OWA operator to resemble the Minimum or Maximum operators. Those operators are numerical implementations of logical operators $AND$ and $OR$. To handle that, the \textit{orness} of the OWA operator was defined by Yager as a measure of how close to the $OR$ operator the OWA is.

A common problem is to determine the vector weight of an OWA operator that meets a desired orness. A lot of methods have been proposed to solve this problem \cite{Liu2011}. In this paper we propose another one. Our method is fast, robust, has good performance and is easy to interpret by non-experts. Some of these features appear in some methods, and other features in others, but our method accomplishes all of them. As an example, consider the exponential method \cite{Yager1998}: it is fast and robust, but its performance (measured with the entropy index) is not the best; our method is faster, as robust and performs better than the exponential. As another example, consider the Maximum Entropy method \cite{Fuller2000}, whose performance is optimal, but it is slow and have numerical problems; our method is faster and more robust than it, and its performance is very close to the optimal.

The structure of the paper is the following: section \ref{sec:owas} summarizes the background about OWA operators. In section \ref{sec:family} we propose a new method to obtain the weight vector of an OWA operator that meets a desired orness. We explore the behaivor and results of the proposed method in section \ref{sec:examples}. Some conclusions are done in section \ref{sec:conclusions}.

\section{OWA operators}
\label{sec:owas}
 OWA operators are aggregation operators defined by \cite{Yager1998,Detyniecki2001}:
 \begin{equation}
  \begin{array}{c}
  OWA: \mathcal{R}^n \to \mathcal{R} \\ \\
  y=f(x_1,x_2,\cdots, x_n) \\ \\
  y=\sum_{i=1}^n w_i \bar{x}_i\qquad \sum_{i=1}^n w_i=1\qquad 0\leq w_i \leq 1\;\;\forall i \\
  \end{array}
 \end{equation}
 
 Where $\bar{x}_i$ is the $i-$th element of $\{x_1,x_2,\cdots,x_n\}$ previously ordered from highest to lowest. The selection of the weight vector $\mathbf{w}=\{w_1,\cdots,w_n\}$ stablishes different kind of aggregations; remarkable OWA operators are:
 \begin{description}
  \item [Maximum:] with $\mathbf{w}=\{1,0,0,\cdots,0\}$
  \item [Minimum:] with $\mathbf{w}=\{0,0,\cdots,0,1\}$
  \item [Simple average:] with  $\mathbf{w}=\{1/n,1/n,\cdots,1/n\}$
 \end{description}

\subsection{Orness and entropy}

Yager define $orness(\mathbf{w})$ and $disp(\mathbf{w})$ (dispersion or entropy) as:
\begin{equation}
 \label{equ:ornessYager}
 orness(\mathbf{w})=\frac{1}{n-1}\sum_{i=1}^n(n-i)w_i
\end{equation}

\begin{equation}
 \label{equ:entropyYager}
 disp(\mathbf{w})=-\sum_{i=1}^n w_i \ln{w_i}
\end{equation}

Orness refers to the degree to which the aggregation is like an \textit{or} operation where as dispersion measures the degree to which $\mathbf{w}$ takes into account all information in the aggregation.

\subsection{Families of OWA operators}
There are a lot of strategies to compute the weight vector of an OWA operator. An extensive review is found in \cite{Liu2011} where they are classified in 5 groups:
\begin{enumerate}[1.]
 \item Optimization methods. The problem is stated as the optimization of a performance index subject to restrictions. Shanon entropy, dispersion, Rényi entropy are some of the indexes used. One of the most remarkable method is the Maximum Entropy method proposed by O'Hagan \cite{Ohagan1988} that was reformulated in \cite{Fuller2000} in an analytical way.
 \item Empirical data methods. This group of methods are usually intended to model the risk attitude of a decision maker: from a set of recorderd decisions we must find the OWA that better reflects them.
 \item Regular Increasing Monotone (RIM) quantifier methods. RIMs are lingüistic quantifiers that can be used to guide the selection of the weight vector.
 \item The argument dependent methods. These are methods that use the input data vector $\mathbf{x}$ as useful information in the weight vetor determination problem. Unlike the classical OWA operators, inputs do not have to be ordered.
 \item Preference relation methods. These methods are an extension of the empirical data methods. Empirical data is replaced by a preference matrix provided by the experts.
\end{enumerate}

A sixth group of methods can be added. It can be called Shaped based methods, because they use a predefined shape of the weight distribution to solve the problem. Notable methods are the exponential proposed by Yager in \cite{Yager1998} and the linear proposed by Lamata in \cite{Lamata2009}.

\section{A family of OWA operators}
\label{sec:family}
We now address the issue to find the weights of an OWA operator with a desired orness. The operator must aggregate the information of $n$ individual values. We define $\alpha \in [0,1]$ as a variable that equals the desired orness. However, in a more general perspective, $\alpha=g(orness) \in [0,1]$ with $g(\cdot)$ a monotonically increasing function with $g(0)=0$ and $g(1)=1$.

We impose the following restrictions, intended to construct a family whose behaivor is easy to understand:
\begin{enumerate}[{OWA.}1 ]
 \item if $\alpha=0.0$ the operator must be equivalent to the Mimimum operator.
 \item if $\alpha=0.5$ the operator must be equivalent to the Simple Average operator.
 \item if $\alpha=1.0$ the operator must be equivalent to the Maximum operator.
 \item \label{item:sim} operator's behaivor must be symetrical respect to $\alpha=0.5$. In other words, the way in which the operator evolves from Minimum to Simple average when $\alpha$ varies from $0.0$ to $0.5$ must be the same way in which it evolves from Maximum to Simple Average when $\alpha$ varies from $1.0$ to $0.5$.
\end{enumerate}

Last restriction is easy to acomplish if we define a family for $0\leq\alpha\leq 0.5$ and then we use simetry to define the behaivor for $0.5 < \alpha \leq 1.0$.

\subsection{And-like operators ($0\leq\alpha\leq0.5$)}
\begin{figure}
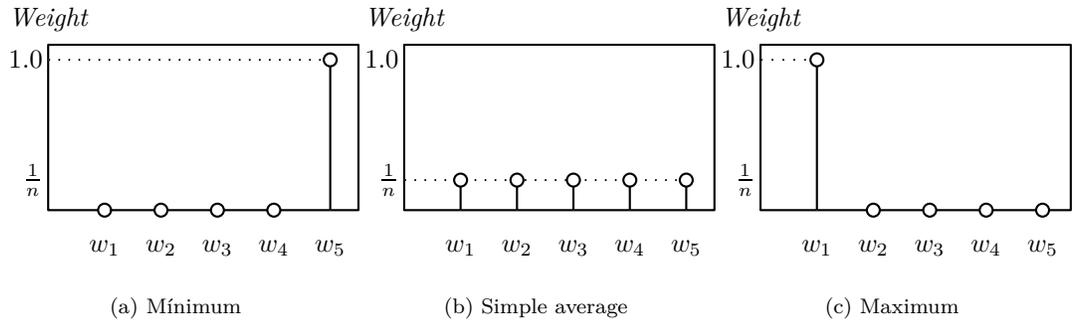

\centering
 \subfloat[Mínimum]       {\label{fig:owaMin} \owasSimple{1}}
 \subfloat[Simple average]{\label{fig:owaMed}\owasSimple{2}}
 \subfloat[Maximum]       {\label{fig:owaMax}\owasSimple{3}}
 \caption{Weigths for 3 OWAs with $n=5$}
 \label{fig:3owas}
\end{figure}

Be $\alpha$ the desired orness, and consider first the case $0\leq\alpha\leq0.5$. The rationale of our proposal is the following: as $\alpha$ varies from $0$ to $0.5$ the weights must vary from those shown in figure~\subref*{fig:owaMin} to those in figure~\subref*{fig:owaMed}. We first vary $w_n$ monotonically from $1.0$ to $1/n$. The amount in which we diminishes $w_n$ is distributed into the other weights to keep unchanged the sum of weights. We distribute this amount using an straight line whose parameters are easy to compute, as we show in the following paragraphs.

We define $f(x):[0, 0.5]\to[0,1]$ a monotonically increasing function with $f(0)=0$ y $f(0.5)=1$. Other conditions for $f(x)$ are developed later. The value of $w_n$  must vary from $1$ to $1/n$ when $\alpha$ varies from $0$ to $0.5$. We use $f(\cdot)$:

\begin{equation}
 \label{equ:w0}
w_n=1-\Delta \qquad \Delta=f(\alpha)(1-1/n)=f(\alpha)(n-1)/n
\end{equation}
Under this conditions, the sum of the other weights must be
\begin{equation}
 \label{equ:sumaD}
 \sum_{i=1}^{n-1}w_i=\sum_{i=1}^{m}w_i=\Delta\qquad m=n-1
\end{equation}
According with equation \ref{equ:ornessYager} the orness is
\begin{equation}
 \label{equ:ornessD}
orness(\mathbf{w})=\frac{1}{n-1}\sum_{i=1}^n(n-i)w_i=\frac{1}{m}\sum_{i=1}^m(n-i)w_i=\alpha
\end{equation}

We propose a distribution of $\Delta$ following the linear relation:
\begin{equation}
 \label{equ:wi}
 w_i=Ki+b\qquad i=1,2,\cdots,m
\end{equation}
Notice that this proposal is different from the linear proposal of Lamata \cite{Lamata2009}, because we use the linear relation for $n-1$ weights and Lamata uses it for all the $n$ weights.

Equations \ref{equ:sumaD} and \ref{equ:ornessD} turns into
\begin{equation}
 \left\{
 \begin{array}{rl}
  \sum_{i=1}^{m}(Ki+b) & =\Delta \\ \\
  \frac{1}{m}\sum_{i=1}^m(n-i)(Ki+b) & =\alpha
 \end{array}
 \right.
\end{equation}

That can be arranged as
\begin{equation}
 \left\{
 \begin{array}{rl}
  K\sum_{i=1}^{m}i +\sum_{i=1}^{m}b & =\Delta \\ \\
  \frac{K}{m}\left(n\sum_{i=1}^{m}i - \sum_{i=1}^{m}i^2\right) + \frac{b}{m}\left(\sum_{i=1}^{m}n - \sum_{i=1}^{m}i\right) & =\alpha
 \end{array}
 \right.
\end{equation}
We compute the  sums using Faulhaber's formulas
\[
 \sum_{i=1}^m i = \frac{m(m+1)}{2}
\qquad \qquad \sum_{i=1}^m i^2 = \frac{m(m+1)(2m+1)}{6}
\]

\begin{equation}
 \left\{
 \begin{array}{rl}
  K\frac{m(m+1)}{2} + mb & =\Delta \\ \\
  \frac{K}{m}\left(\frac{m(m+1)^2}{2} - \frac{m(m+1)(2m+1)}{6}\right) + \frac{b}{m}\left(m(m+1) -\frac{m(m+1)}{2} \right) & =\alpha
 \end{array}
 \right.
\end{equation}

And rearrange the terms in order to show explicitly a system of 2 linear equations and 2 unknows
\begin{equation}
 \label{equ:sistema}
 \left\{
 \begin{array}{rcrl}
  K\frac{m(m+1)}{2} & + & bm & =\Delta \\ \\
  K\frac{(m+1)(m+2)}{6} & + & b\frac{m+1}{2} & =\alpha
 \end{array}
 \right.
\end{equation}

From \ref{equ:sistema} we get direct expressions for $K$ and $b$
\begin{equation}
 \label{equ:KbDelta}
 \left\{
 \begin{array}{rl}
  K = & 6\frac{\Delta n/m -2\alpha}{m^2 - 1}\\ \\
  b = & \frac{\Delta}{m} - K\frac{(m+1)}{2}
 \end{array}
 \right. 
\end{equation}

Using equation \ref{equ:w0} we can compute $K$ and $b$ from $\alpha$ and $m$ or $n$
\begin{equation}
 \label{equ:KbAlpha}
 \left\{
 \begin{array}{rlcl}
  K = & 6\frac{f(\alpha) -2\alpha}{m^2 - 1} & = & 6\frac{f(\alpha) -2\alpha}{n(n-2)} \\ \\
  b = & \frac{f(\alpha)}{m+1} - K\frac{(m+1)}{2} & = & \frac{f(\alpha)}{n} - K\frac{n}{2} 
 \end{array}
 \right. 
\end{equation}

\subsection{Selection of $f(\cdot)$}
As all weights must be non-negative, lets consider the first weight $w_1= K\times 1+b$
\begin{equation}
 \begin{array}{rl}
 K+b & \geq 0 \\ \\
 K + \frac{\Delta}{m} - K\frac{m+1}{2} & \geq 0 \\ \\
 K\frac{1-m}{2} + \frac{\Delta}{m} & \geq 0 \\ \\
 K\frac{1-m}{2}  & \geq -\frac{\Delta}{m} \\ \\
 3(1-m)\frac{\Delta n/m -2\alpha}{m^2-1}& \geq -\frac{\Delta}{m} \\ \\
 -\frac{3}{n}(\Delta\frac{n}{m} - 2\alpha) & \geq -\frac{\Delta}{m} \\ \\
 2m\alpha - n\Delta  & \geq -n\Delta/3 \\ \\
 2m\alpha  & \geq 2n\Delta/3 \\ \\
 3\alpha & \geq \Delta n/m \\ \\
 3\alpha & \geq f(\alpha) \\ \\
 \end{array}
\end{equation}

To ensure that the other weights are also non-negative, we impose the condition $K\geq 0$. Directly from equation \ref{equ:KbAlpha} we get $f(\alpha) \geq 2\alpha$. 

We have found the following restrictions for $f(\alpha)$:
\begin{equation}
 \label{equ:limitsF}
 2\alpha \leq f(\alpha) \leq 3\alpha
\end{equation}

\begin{figure}
 \centering
  \begin{pspicture}(0,0)(8,5)
  \rput[bl](1,1)
  {
    \psset{xunit=10}
    \psset{yunit=3}
    \psaxes[Dx=0.5]{->}(0,0)(0.6,1.1)[$\alpha$,0][$f(\alpha)$,90]
    \psline[linestyle=dashed](0,0)(0.5,1)
    \psline[linestyle=dashed](0,0)(0.333,1)
    \rput[t]{40}(0.25,0.48){$2\alpha$}
    \rput[b]{45}(0.25,0.78){$3\alpha$}
    \psplot[plotpoints=20,linewidth=2pt]{0}{0.5}{x -2 mul 1 add 1.25 exp -1 mul 1 add}
  }
 \end{pspicture}
 \caption{Function $f(\alpha)= 1 - (2\alpha - 1)^\beta$ and its restrictions}
 \label{fig:falfa}
\end{figure}
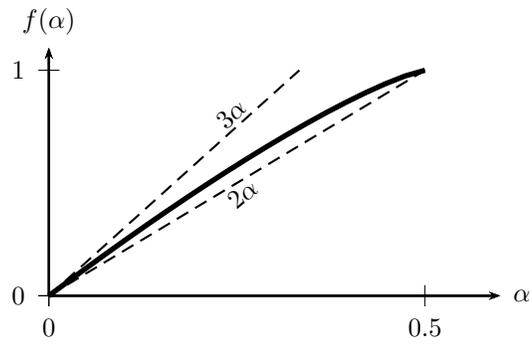

If we choose $f(\alpha)= 1 - (1-2\alpha)^\beta$ with some $0 \leq \beta \leq 1$ (figure \ref{fig:falfa}), those restrictions are satisfied if the slope at $0$ is in $[2,3]$
\begin{equation}
 \begin{array}{rcl}
  2 \leq & \frac{df(\alpha)}{d\alpha}\vert_{\alpha=0} & \leq 3 \\ \\
  2 \leq & 2\beta(1-2\alpha)^{\beta-1}\vert_{\alpha=0} & \leq 3 \\ \\
  2 \leq & 2\beta & \leq 3 \\ \\
  1.0 \leq & \beta & \leq 1.5 \\ \\
 \end{array}
\end{equation}

Notice that the limit case with $\beta=1.0$ makes $f(\alpha)=2\alpha$. It makes $K=0$ and $b=1/n$ meaning that $\Delta$ is homogeneously distributed between $w_1,\cdots,w_{n-1}$.

\subsection{Or-like operators ($0.5\leq\alpha\leq1$)}

To ensure a symetrical behaivor of the family, we need to adapt the previous result when  $0.5\leq\alpha\leq 1.0$ in the following way:
\begin{itemize}
 \item Compute the weights for $\bar{\alpha} = 1 - \alpha$
 \item Reverse the order of the weights.
\end{itemize}

Algorithm \ref{alg:linear} implements the above results in pseudocode.

\section{Examples and analysis}
\label{sec:examples}
In order to analyze the algorithm performance, we compare their results for 3 values of $\beta$ ($1.0$, $1.25$ and $1.5$) against two well known algorithms: exponential method \cite{Yager1998} and maximum entropy method \cite{Fuller2000}. Exponential method is similar to our proposal in the sense that it also propose a shape for the distribution of $\Delta$ (an exponential distribution); however, to fit the desired orness, $\alpha$ must be preset according to pre-computed figures. Maximum entropy method has an optimization approach, and it involves the numerical solution of an equation.

\begin{figure}
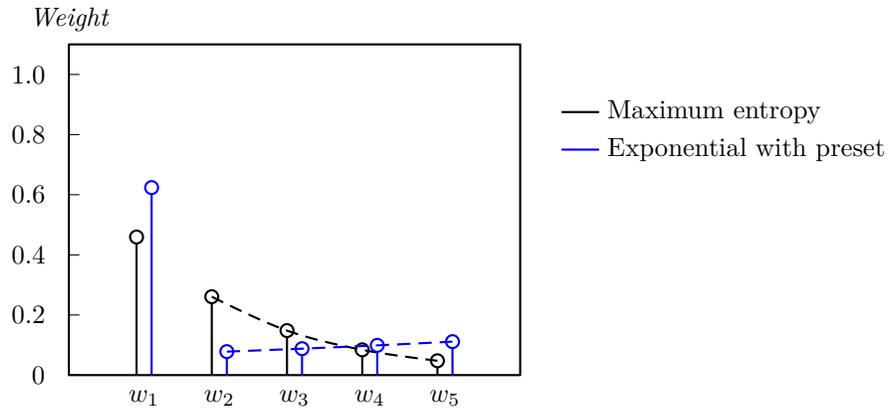
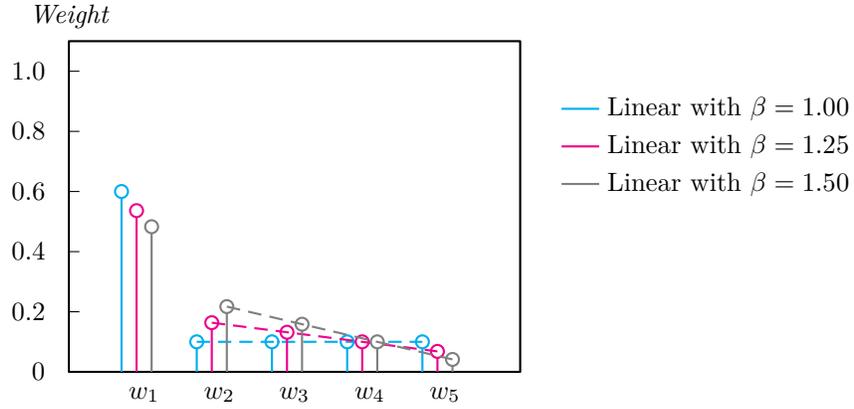

 \centering
 \subfloat[Exponential and Entropy methods]{\label{fig:weightsComparisonA}\plotweights{1}{A}} \\
 \subfloat[Linear method]                  {\label{fig:weightsComparisonB}\plotweights{1}{B}} \\ 
 \caption{Weights for $n=5$ and desired orness $0.6$.}
 \label{fig:weightsComparison}
\end{figure}

Figure \ref{fig:weightsComparison} shows the weights obtained for $n=5$ when the desired orness is $0.6$. Notice the linear distribution from $w_2$ to $w_5$ in figure~\subref*{fig:weightsComparisonA} and the Homogeneous distribution when $\beta=1.0$. Notice also that the linear method approaches better than the exponential the weight distribution of the optimal entropy method. In fact, the exponential method provides an increasing distribution where as entropy and linear methods provide decreasing distributions.

\begin{figure}
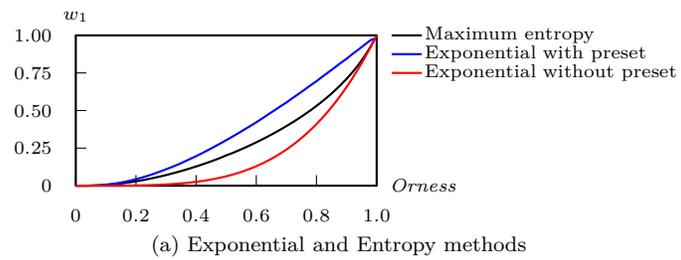
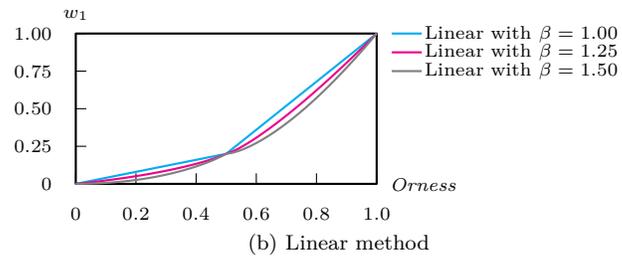

\centering
 \subfloat[Exponential and Entropy methods]{\label{fig:w1ComparisonA}\variableVsOrness{W1}{$w_1$}{1}{2}{B}} \\
 \subfloat[Linear method]                  {\label{fig:w1ComparisonB}\variableVsOrness{W1}{$w_1$}{1}{2}{C}} \\
 \caption{Variation of $w_1$ for $n=5$ as a function of the desired orness}
 \label{fig:w1Comparison}
\end{figure}

Figure \ref{fig:w1Comparison} shows how fast varies $w_1$ when we change the desired orness for $n=5$. We have added the case of the exponential method without preset, in order to stress the presetting effect. Curves in figure~\subref*{fig:w1ComparisonB} has a change when orness is $0.5$, caused by the strategy adopted to ensure simetry. These figures also show that linear method behaves closer to entropy method than the exponential one. 

\begin{figure}
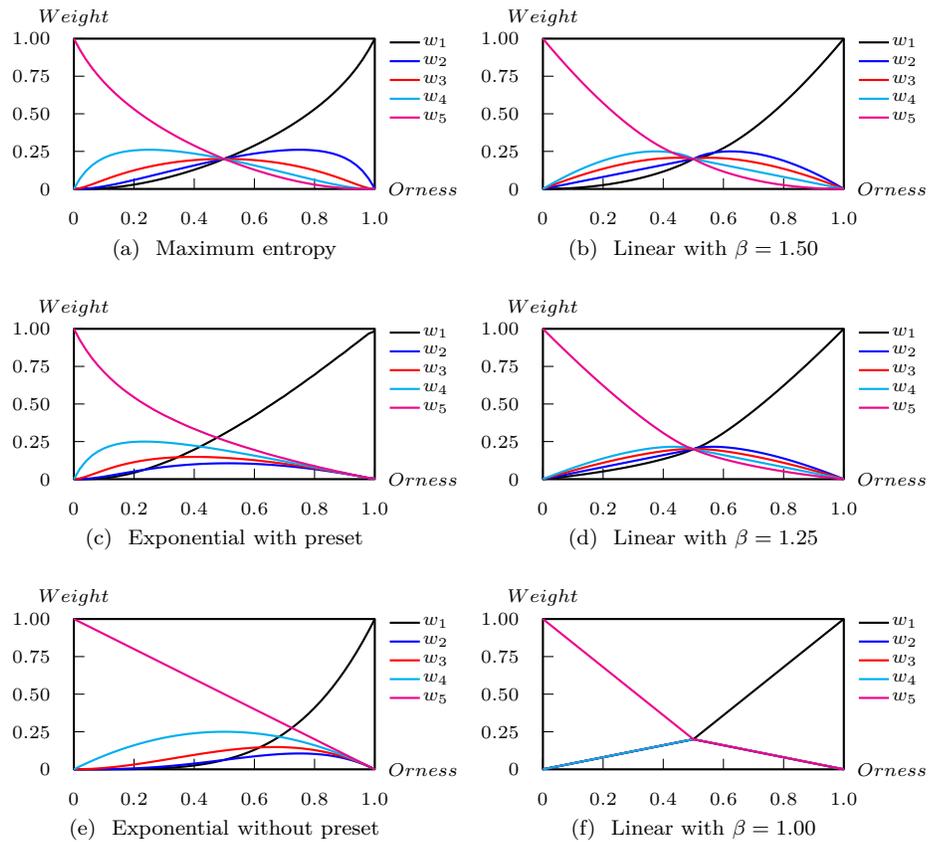

\centering
 \subfloat[ \protect\labelcaso{1}]{\label{fig:weightsComparisonDuarteA}\weightVsOrness{1}}
 \subfloat[ \protect\labelcaso{6}]{\label{fig:weightsComparisonDuarteB}\weightVsOrness{6}}\\
 \subfloat[ \protect\labelcaso{2}]{\label{fig:weightsComparisonDuarteC}\weightVsOrness{2}}
 \subfloat[ \protect\labelcaso{5}]{\label{fig:weightsComparisonDuarteD}\weightVsOrness{5}}\\
 \subfloat[ \protect\labelcaso{3}]{\label{fig:weightsComparisonDuarteE}\weightVsOrness{3}}
 \subfloat[ \protect\labelcaso{4}]{\label{fig:weightsComparisonDuarteF}\weightVsOrness{4}}
 \caption{Comparison of Weights for $n=5$ as a function of desired Orness}
 \label{fig:weightsComparisonDuarte}
\end{figure}

Figure \ref{fig:weightsComparisonDuarte} shows the change in all the weights when we change the desired orness for $n=5$. It is important to remark the simetry of the entropy and linear methods. Similarity between figures \subref*{fig:weightsComparisonDuarteA} and \subref*{fig:weightsComparisonDuarteB} reinforces the idea that linear method with $\beta=1.5$ is very similar to maximum entropy method.

\begin{figure}
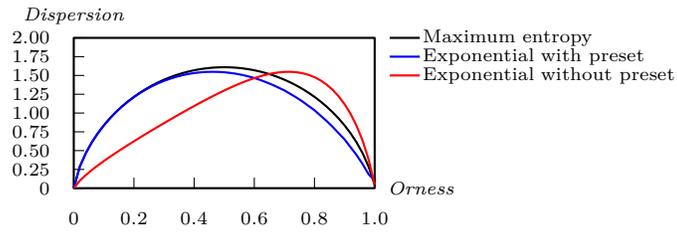
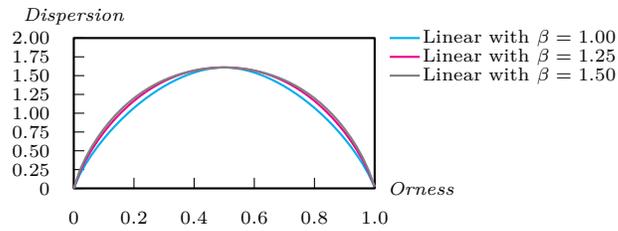
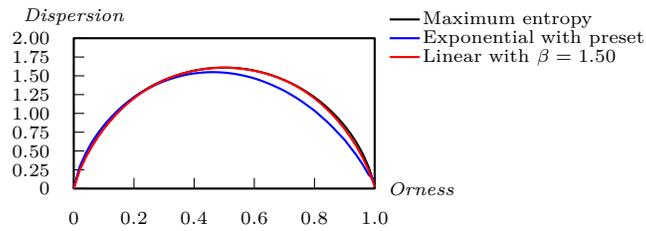

\centering
 \subfloat[Exponential and Entropy methods]%
 {\label{fig:comparisonEntropyA}\variableVsOrness{DISPERSION}{Dispersion} {2}{1}{B}} \\
 \subfloat[Linear method]%
 {\label{fig:comparisonEntropyB}\variableVsOrness{DISPERSION}{Dispersion} {2}{1}{C}} \\
 \subfloat[Exponential, Entropy and Linear method]%
 {\label{fig:comparisonEntropyC}\variableVsOrness{DISPERSION}{Dispersion} {2}{1}{D}} \\
 \caption{Comparison of Orness and Dispersion (Entropy) for $n=5$ as a function of desired Orness}
 \label{fig:comparisonEntropy}
\end{figure}

Figure \ref{fig:comparisonEntropy} shows how varies dispersion (entropy) when we change the desired orness for $n=5$. Variations of entropy with the linear method for different $\beta$ are minimal. Notice also that the entropy obtained by the linear method with $\beta=1.5$ is very close to the optimal obtained by the entropy method.

Table~\ref{tab:times} shows a comparison of computing times. Algorithms have been implemented in OpenModelica \cite{Fritzson2014}. We have run every algorithm 20 times for $n=10$ and $n=100$. The average simulation times are in table \ref{tab:times}, as relative times to the minimal average time. Exponential method, as described in \cite{Yager1998}, involves a graphical procedure for the preset of $\alpha$; we have implemented it using an exhaustive search procedure, which explains the high times for this method. It is an unfair comparison, and we prefer to compare exponentials method times using an implementation without preseting $\alpha$.

Table~\ref{tab:times} shows that the linear method is the fastest one. It is faster than the entropy method, because it doesn't need any iteration loop and it is faster than the exponential method because arithmetic operations are simpler.

\begin{figure}
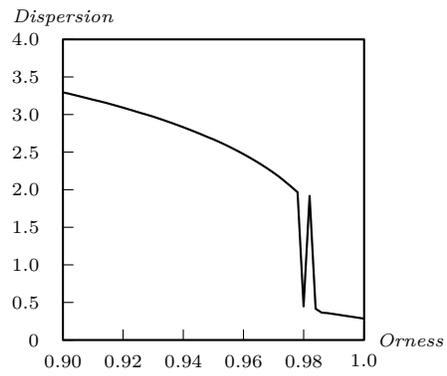
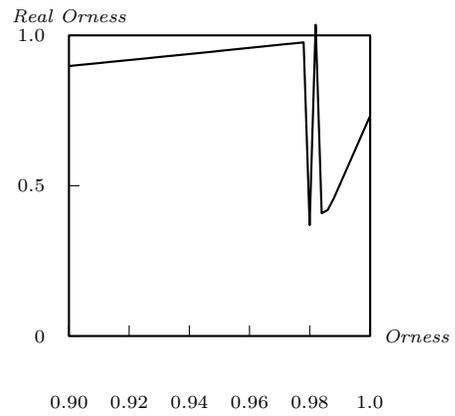

\centering
 \subfloat[Dispersion]%
  {\variableVsOrnessNcien{DISPERSION}{Dispersion} {4}{1}}
 \subfloat[Real orness]%
  {\variableVsOrnessNcien{ORNESS}{Real Orness} {1}{4}}
 \caption{Dispersion (Entropy) and Real Orness for the Maximum entropy method with desired orness close to Maximum, for $n=100$}
 \label{fig:entropyN100}
\end{figure}

Moreover, entropy method has important lacks (see remark 1 in \cite{Liu2011}): objective function definition implies $w_i>0$ and as Maximum and Minimum operators are OWA's with most of the weights equal to zero, those cases can not be included. Even for $\alpha$ close to $0$ or $1$ numerical problems may arise: Figure~\ref{fig:entropyN100} shows the dispersion (entropy) and orness for the Maximum entropy method for desired orness $\in [0.9,1.0]$ obtained with our OpenModelica implementation. Notice the numerical problems thah appear near $\alpha =0.98$. Those numerical problems are not present in the linear method, because there are not iteration loops.

\section{Conclusions}
\label{sec:conclusions}
The linear method is a fast and reliable method for computing weights of OWA operators with a desired orness. It is as fast or faster than the exponential method. It is also more suitable for software implementations than the exponential method because does not use precomputed curves for presettings. It is more robust than optimization methods because there are no iterations loops. And it provides a suboptimal result, very close to the optimal.

\section*{References}
 \bibliographystyle{elsarticle-harv}
 \bibliography{bib/owa}
 
\newpage

\begin{algorithm}[H]
 \SetKwInOut{Input}{input}\SetKwInOut{Output}{output}
 \Input{$n$ the size of the OWA operator}
 \Input{$orness$ the desired orness of the OWA operator}
 \Output{$w$ a vector of size $n$ with the weights of the OWA operator\\\rule{2cm}{0cm}}
 
 Define $f(x)=1 - (1-2x)^\beta$ with $\beta \in [1,1.5]$. By default $\beta=1.5$\\
 \eIf{$orness \leq 0.5$}{$\alpha \leftarrow orness$\\}{$\alpha \leftarrow 1 - orness$}
 $m \leftarrow n-1$\\
 $\Delta \leftarrow f(\alpha)m/n$\\
 $K \leftarrow 6(f(\alpha) - 2\alpha)/(m^2-1)$\\
 $b \leftarrow f(\alpha)/n - Kn/2$\\
 \eIf{$orness \leq 0.5$}
 {
  $w_n \leftarrow 1 - \Delta$\\
  \For{$i\leftarrow 1$ \KwTo $m$}
  {
    $w_i \leftarrow Ki+b$
  }
 }
 {
  $w_1 \leftarrow 1-\Delta$\\
  \For{$i\leftarrow 1$ \KwTo $m$}
  {
    $w_{i+1} \leftarrow K(n-i)+b$
  }
 }
 \centering
 \caption{Algorithm to get the weights of an OWA operator using the linear method}
 \label{alg:linear}
\end{algorithm}

\newpage

\newpage
\clearpage

\begin{table}
\centering
\caption{Relative computing times. Averages from 20 samples}
\label{tab:times}
\begin{tabular}{lcc}\hline
 \textbf{Method} & $n=10$ & $n=100$ \\ \hline \hline
 \labelcaso{1} & 2.91 & 2.26 \\ \hline
 \labelcaso{2} & 108.81 & 366.22 \\ \hline
 \labelcaso{3} & 1.05 & 2.04 \\ \hline
 \labelcaso{4} & 1.05 & 1.01 \\ \hline
 \labelcaso{5} & 1.03 & 1.02 \\ \hline
 \labelcaso{6} & 1.00 & 1.00 \\ \hline
\end{tabular}

\end{table}

\end{document}